%% file: main.tex
\def\usearxivstyle{1}
\ifdefined\usearxivstyle

\documentclass[11pt]{article}
\usepackage[numbers]{natbib}
\usepackage{packages}
\usepackage{statistics}
\usepackage{statistics-macros}

\usepackage{manyfoot}

\definecolor{innerboxcolor}{rgb}{.9,.95,1}
\definecolor{outerlinecolor}{rgb}{.6,0,.2}
\definecolor{outerlinecoloreb}{rgb}{0,1,0}
\definecolor{innerboxcoloreb}{rgb}{1,1,1}

\newcommand{\dottedline}[1]{%
  \multicolumn{1}{c}{\dotfill} & \multicolumn{1}{c}{\dotfill}\\
}

\DeclareNewFootnote{A}
\DeclareNewFootnote{B}

\let\footnoteR\footnoteB
\let\footnote\footnoteA

\begin{document}


\begin{center}
  {\huge Topology-Aware Dynamic Reweighting for Distribution Shifts on Graph} \\
  \vspace{.5cm}
  {\large
    Weihuang Zheng\footnoteR{Equal Contributions}$^{,1}$, Jiashuo Liu\textsuperscript{*}$^{,2}$, Jiaxing Li$^{1}$, Jiayun Wu$^{2}$, Peng Cui$^{2}$, Youyong Kong$^1$} \\
  \vspace{.2cm}
   $^1$ School of Computer Science and Engineering, Southeast University\\
   $^2$Department of Computer Science and Technology, Tsinghua University\\
    \vspace{.2cm}
  \texttt{zhengweihuang@seu.edu.cn, liujiashuo77@gmail.com}\\
  \texttt{cuip@tsinghua.edu.cn, kongyouyong@seu.edu.cn}
\end{center}


\else

\documentclass[twoside]{article}

\usepackage{aistats2024}
\usepackage{natbib}
\usepackage{amsmath}
\usepackage{amssymb}
\usepackage{mathtools}
\usepackage{amsthm}
\usepackage{xcolor}   
\usepackage{mathrsfs}
\usepackage{algorithm}
\usepackage{algorithmic}
\usepackage{graphicx}
\usepackage{subfig}
\usepackage{bbding}
\usepackage{multirow}
\usepackage{tabulary}
\usepackage{colortbl}
\usepackage{color}
\usepackage[utf8]{inputenc} 
\usepackage[T1]{fontenc}    
\usepackage{hyperref}       
\usepackage{url}            
\usepackage{booktabs}       
\usepackage{amsfonts}       
\usepackage{nicefrac}       
\usepackage{microtype}      
\usepackage{xcolor}         
\usepackage{amsmath}
\usepackage{amsthm}
\usepackage{amssymb}
\usepackage{mathrsfs}
\usepackage{algorithm}
\usepackage{algorithmic}
\usepackage{graphicx}
\usepackage{bbding}
\usepackage{multirow}
\usepackage{mathtools,color}
\newtheorem{assumption}{Assumption}[section]
\newtheorem{definition}{Definition}[section]
\newtheorem{proposition}{Proposition}[section]
\newtheorem{theorem}{Theorem}[section]
\newtheorem{lemma}{Lemma}[section]
\newtheorem{problem}{Problem}
\newtheorem*{example}{Example}
\newtheorem*{insight}{Insight}
\newtheorem*{remark}{Remark}
\newcommand{\tabincell}[2]{\begin{tabular}{@{}#1@{}}#2\end{tabular}}

\definecolor{babypink}{rgb}{0.96, 0.76, 0.76}
\definecolor{bittersweet}{rgb}{1.0, 0.44, 0.37}
\definecolor{blizzardblue}{rgb}{0.67, 0.9, 0.93}
\definecolor{brickred}{rgb}{0.8, 0.25, 0.33}
\definecolor{bubbles}{rgb}{0.91, 1.0, 1.0}
\definecolor{classicrose}{rgb}{0.98, 0.8, 0.91}
\definecolor{languidlavender}{rgb}{0.84, 0.79, 0.87}
\definecolor{pastelred}{rgb}{1.0, 0.41, 0.38}
\definecolor{lightpink}{rgb}{1.0, 0.71, 0.76}
%
%




\begin{document}

%

%

\twocolumn[

\fi

\begin{abstract}
Graph Neural Networks (GNNs) are widely used for node classification tasks but often fail to generalize when training and test nodes come from different distributions, limiting their practicality. 
To overcome this, recent approaches adopt invariant learning techniques from the out-of-distribution (OOD) generalization field, which seek to establish stable prediction methods across environments. 
However, the applicability of these invariant assumptions to graph data remains unverified, and such methods often lack solid theoretical support. 
In this work, we introduce the \textbf{T}opology-\textbf{A}ware Dynamic \textbf{R}eweighting (TAR) framework, which dynamically adjusts sample weights through gradient flow in the geometric Wasserstein space during training. 
Instead of relying on strict invariance assumptions, we prove that our method is able to provide distributional robustness, thereby enhancing the out-of-distribution generalization performance on graph data.
By leveraging the inherent graph structure, TAR effectively addresses distribution shifts.
Our framework's superiority is demonstrated through standard testing on four graph OOD datasets and three class-imbalanced node classification datasets, exhibiting marked improvements over existing methods.
\end{abstract}

\input{./data/1intro}

\input{./data/2pre}

\input{./data/3method}

\input{./data/4exp}
\input{./data/5con}

\newpage
\bibliography{reference.bib}

 \def\usearxivstyle{1}
\ifdefined\usearxivstyle
\bibliographystyle{abbrvnat}
\else
\bibliographystyle{apalike}
\fi

\newpage
\input{data/6appendix.tex}

\end{document}

%% file: data/1intro.tex
\section{Introduction}


Graph Neural Networks (GNNs) have been widely used in node classification tasks, such as advertising recommendation \cite{jiang2023adaptive}, social network anomaly detection \cite{tang2022rethinking}, etc. 
However, these GNN models typically assume that the training and test graph data are drawn from the same distribution, which does not always hold in practice. 
In real-world graph data, sample selection bias~\cite{fan2022debiased,he2020learning} as well as graph construction techniques~\cite{qiao2018data,zhou2023opengsl} often brings distribution shifts between training nodes and test nodes.
For instance, In WebKB~\cite{pei2020geom} datasets, web pages (nodes) and categories (labels) are heavily affected by the university they originate from, leading to distribution shifts among nodes drawn from different universities. 
Therefore, in order to enhance the practical validity of GNNs, it is of paramount importance to deal with distribution shifts on graph data.

To address the distribution shift problem in node classification, recent works~\cite{li2022learning,EERM,sui2023unleashing,wu2022discovering,liu2023flood} borrow the idea of invariant learning methods from the literature of out-of-distribution (OOD) generalization and adopt them on graph-structured data.
Invariant learning \cite{arjovsky2019invariant,liu2021heterogeneous} stems from the causal inference literature, and now becomes one of the key approaches to solving OOD problems on graphs. 
The core concept is to identify invariant features with stable prediction mechanisms across different environments, thereby mitigating performance degradation under distribution shifts. 
And most of the works in this line directly apply existing invariant learning algorithms to graph-level classification tasks (\emph{major})~\cite{li2022learning,sui2023unleashing,liu2023flood,yu2023mind} and node classification tasks (\emph{minor})~\cite{EERM,xia2024learning}.
However, methods based on invariance learning are built upon strong invariance assumptions that lack further validation for their actual validity~\cite{liu2024need}.
And there also lack guarantees regarding whether the invariant representations are truly learned on complex graph-structured data. 
Besides, sample reweighting methods are also utilized to handle distribution shifts in node classifications~\cite{gui2022good}, typified by Group DRO~\cite{sagawa2019distributionally}, while they ignore the complex topological structure information of the graph data, and the reweighting scheme relies on the pre-defined subgroups.

In this work, we focus on the distribution shift problem on node classification tasks, and propose the \textbf{T}opology-\textbf{A}ware Dynamic \textbf{R}eweighting (TAR) framework to enhance the generalization ability of GNN models.
Our TAR framework involves a minimax procedure, where the inner maximization problem learns sample probability densities under the entropy and topology constraints, and the outer minimization problem optimizes the GNN model under the learned distribution.
For the reweighting scheme (inner problem), as demonstrated in Section~\ref{sec:optimization}, we perform gradient flow in a new metric space, named geometric Wasserstein space, where the distance metric is the optimal transport along the graph structure.
In this way, we	\emph{incorporate the topological structure information} into the learning of sample probability densities, and the change of sample weights is restricted on graph edges (as shown in Equation~\ref{equ:gradient}).
Furthermore, in Section~\ref{sec:theoretical}, we prove that our gradient flow procedure is equivalent to finding the local worst-case distribution, which enhances the distributional robustness of our GNN model. 
We also characterize the error rate introduced by our gradient flow as $e^{-CT_{\text{in}}}$ ($T_{\text{in}}$ is the number of steps).
Finally, experimental results on \textbf{4} typical OOD and \textbf{3} class-imbalanced node classification datasets demonstrate the effectiveness of our proposed TAR framework.

%% file: data/2pre.tex
\section{Preliminaries}

\textbf{Notations.}\quad 
$X\in\mathcal{X}$ denotes the covariates, $Y\in\mathcal{Y}$ denotes the target, $\mathbb P_{s}(X,Y)$ and $\mathbb P_{t}(X,Y)$ represent the joint source distribution and the target distribution, abbreviated with $\mathbb P_{s}$ and $\mathbb P_t$ respectively.
The prediction model is denoted by $f_\theta(\cdot):\mathcal{X}\rightarrow\mathcal{Y}$, for which we use graph neural networks (GNN) throughout this paper.
$[N]=\{1,2,\dots, N\}$ denotes the set of integers from 1 to $N$.
The random variable of data points is denoted by $Z=(X,Y)\in\mathcal{Z}$.
A weighted finite graph is denoted by $G_0=(V,E,W)$, where $V=\{v_1,\dots,v_N\}$ is the node set, $E$ is the edge set, and $W=(w_{ij})_{(i,j)\in E}$ are the edge weights. 
$\mathcal{N}(i)$ denotes the set of adjacent nodes for the $i$-th node.

\noindent\textbf{Problem setting.} \quad This work focuses on node classification tasks, where each node in the graph has $d$-dimensional features $X\in\mathbb R^d$, and the task is to predict the class label $Y \in \{1, \ldots, C\}$ via the node's feature (and the graph structure).
Based on this, we define the distribution shift problem in the node classification task. 
The joint data distribution can be decomposed as $\mathbb P(Y, X) =\mathbb P(Y|X)\mathbb P(X)$. The main causes of distribution shifts can be separated into two types of shifts:
(1) Covariate shift (\(\mathbb P_s(Y|X) = \mathbb P_t(Y|X), \mathbb P_s(X) \neq \mathbb P_t(X) \)): This indicates that the feature distribution differs between the source and the target.
(2) Concept shift (\(\mathbb P_s(Y|X) \neq \mathbb P_t(Y|X),\mathbb P_s(X) = \mathbb P_t(X) \)): This indicates that there are spurious statistical correlations in the source data that may not hold in the target data. 
 Note that we use GNN models throughout this paper, which can be formulated by two step: message aggregation as $\text{agg}(\cdot)$ and representation update as $\text{upd}(\cdot)$. The representation for the $i$-th node at layer $l$ + 1 is defined as:
\begin{equation*}
h_i^{(l+1)} = \text{upd}\left(h_i^{(l)}, \text{agg}({h_j^{(l)} | j \in \mathcal{N}(i)})\right).
\end{equation*}
The overall goal of this work is to \emph{enhance the generalization ability of GNN models} on the node classification tasks under distribution shifts.

In order to mitigate the distribution shift problem on graph data, there are mainly two branches of methods, namely invariant learning~\cite{li2022learning,EERM,SRGNN} and sample reweighting~\citep{sagawa2019distributionally}.
Invariant learning methods rely on the invariance assumption, and propose to identify invariant features across different environments.
However, in node classification tasks, the environments are hard to pre-define, and the actual validity of the invariance assumption itself remains unclear~\cite{liu2024need}, leading to a lack of theoretical guarantees.
Besides, for sample reweighting methods, previous works simply apply Group DRO~\cite{sagawa2019distributionally} on graph data, which ignores the graph structure and treats nodes as independent data points.

In this study, recognizing the limitations of existing approaches, we leverage the inherent topological properties of graph data. 
We propose the Topology-Aware Dynamic Reweighting (TAR) scheme to tackle distribution shift issues in node classification tasks. 
Unlike conventional sample reweighting techniques, TAR relies entirely on the graph structure while incurring only small computational overhead.
Before moving on to our main method, we first provide some preliminaries on the discrete geometric Wasserstein distance, which we use as the topology penalty in our framework (see Equation~\ref{equ:obj} in Section~\ref{sec:method}).

\noindent\textbf{Discrete geometric Wasserstein distance.}\quad We briefly review some key concepts and introduce the discrete geometric Wasserstein distance~\citep{chow2017entropy}, where we adopt the notations used in ~\citep{chow2017entropy, DBLP:conf/nips/LiuW0022}.

The (empirical) probability set supported on all nodes of $G_0$ is denoted as:
\begin{equation*}
	\mathcal P(G_0) = \left\{p = (p_i)_{i=1}^N: \sum_{i=1}^N p_i = 1, p_i\geq 0, \text{for } i\in [N]\right\},
\end{equation*}
which contains all empirical distributions on the node set $V$, and the interior of $\mathcal P(G_0)$ is denoted as $\mathcal P_o(G_0)$.
A \emph{velocity} field $v=(v_{ij})_{i,j\in V}\in \mathbb{R}^{N\times N}$ on graph $G_0$ is a skew-symmetric matrix on the edge set $E$:
\begin{align*}
	v_{ij} = \begin{cases}
		-v_{ij} \quad &\text{if }(i,j)\in E,\\
		 0 \quad &\text{otherwise}.
	\end{cases}
\end{align*}
Given the probability function $p\in\mathcal P(G_0)$ and a velocity field $v$, the \emph{flux} function is defined as the product $pv\in\mathbb R^{N\times N}$:
\begin{equation*}
	pv := (v_{ij}\xi_{ij}(p))_{(i,j)\in E},
\end{equation*}
where $\xi_{ij}(p)$ is a predefined "cross-sectional area", typically interpolated with the associated nodes' densities $p_i,p_j$. 
To ensure the positiveness of $p$ during optimization, we adopt the upwind interpolation from statistical mechanics~\citep{doi:10.13182/NSE81-A20112}: $\xi_{ij}(p)=\mathbb I(v_{ij}>0)p_j + \mathbb I(v_{ij}\leq 0)p_i$ throughout this paper, which relies on the corresponding velocity field.
Intuitively, this characterizes the ``flux'' of sample density from node $i$ to $j$.
Based on this, the \emph{divergence} vector of $pv$ on graph $G_0$ is defined as:
\begin{equation*}
	\text{div}_{G_0}(pv) := -(\sum_{j\in V: (i,j)\in E}\sqrt{w_{ij}}v_{ij}\xi_{ij}(p))_{i=1}^N \in \mathbb R^N,
\end{equation*}
which is supposed to lie in the tangent space of $\mathscr{P}_o(G_0)$.
Intuitively, the $i$-th element in $\text{div}_{G_0}(pv)$ sums over all the in-fluxes and out-fluxes along edges to a certain target node $i$, with each source edge $j$ transporting a probability density $\sqrt{w_{ij}}v_{ij}\xi_{ij}(p)$.

Now we are ready to define the discrete geometric Wasserstein distance:

\begin{definition}[Discrete Geometric Wasserstein Distance~\citep{chow2017entropy}]
\label{def:gw-distance}
	Given a finite graph $G_0$, for any pair of distributions $p^0, p^1 \in \mathscr{P}_o(G_0)$, the discrete geometric Wasserstein distance is defined as:
	\begin{equation*}
	\label{equ:gw}
		\mathcal{GW}_{G_0}^2(p^0,p^1):=\inf\limits_v\left\{\int_0^1 \frac{1}{2}\sum_{(i,j)\in E}\xi_{ij}(p(t))v_{ij}^2 dt: \frac{dp}{dt}+\text{div}_{G_0}(pv)=0, p(0)=p^0, p(1)=p^1\right\},
	\end{equation*}	
	where the infimum is taken over all velocity fields on $G_0$, and $\xi_{ij}(p)$ is a pre-defined interpolation function between $p_i$ and $p_j$.
	Note that $p(t)$ is a continuously differentiable curve $p(t):[0,1]\rightarrow \mathscr{P}_o(G_0)$, which characterizes the probability densities at time $t$.
\end{definition}

\begin{remark}
	In contrast with the conventional Wasserstein distance defined within Euclidean space, the geometric Wasserstein distance necessitates that the transportation of probability density is along the geodesic determined by the graph structure $G_0$.
	In particular, the constraint $\frac{dp}{dt}+\text{div}_{G_0}(pv)=0$ imposes the condition that the change in probability density remains continuous with respect to $G_0$.
\end{remark}

%% file: data/3method.tex
\section{Method}
\label{sec:method}
Motivated by the discrete geometric Wasserstein distance in Definition~\ref{def:gw-distance}, we propose the Topology-Aware Dynamic Reweighting (TAR) algorithm to deal with graph-domain distribution shifts. 

Consider source data $D_s=\{(x_i,y_i)\}_{i=1}^N$ and the corresponding graph structure $G_0=(V,E,W)$.
Denote the empirical marginal distribution as $\hat{\mathbb P}_s$, the overall objective of our TAR algorithm is formulated as:
\begin{equation}
\label{equ:obj}
    \min_{\theta\in\Theta} \max_{q\in\mathcal{P}_o(G_0)} \underbrace{\sum_{i=1}^N q_i\ell(f_\theta(x_i),y_i)}_{\text{Weighted loss}} - \underbrace{\beta\cdot\sum_{i=1}^Nq_i\log q_i}_{\substack{\text{Entropy penalty}}} - \underbrace{\lambda\cdot\vphantom{\sum_i^N}\mathcal{GW}_{G_0}^2(\hat{\mathbb{P}}_s, q)}_{\substack{\text{Topology penalty}}},
\end{equation}
where $\beta$ is the hyper-parameter, and the objective function in general is a minimax optimization over model parameters $\theta$ and sample probability densities $q$.
Note that for the parameter $\lambda$, we set it as $\lambda=\frac{1}{2\tau}$ in our optimization (for details, please refer to Section~\ref{sec:optimization}).
During training, the inner maximization assigns more densities to high-risk samples, thereby prompting the prediction model to prioritize these points. 
This approach aims for a uniformly robust performance across all samples on the graph and helps mitigate potential distribution shifts.
Moreover, to mitigate the risk of overemphasizing unrealistic distributions (e.g., noisy nodes accumulating excessive densities), we introduce entropy and topology penalties as regularization terms.
These penalties integrate topology information for smooth sample weight assignments along the graph structure.

\noindent\textbf{Illustrations.}\quad 
Here we make some remarks on our objective function:\\
(a) \emph{Entropy penalty}: $(-\sum_{i=1}^N q_i\log q_i)$ represents the entropy of empirical probability distribution $q$.
As illustrated in Section \ref{sec:theoretical}, this term serves as a non-linear graph Laplacian operator that encourages sample weights to be smooth along the manifold, avoiding extreme sample weights in the weighted distribution.\\
(b) \emph{Topology penalty}: $\mathcal{GW}_{G_0}^2(\hat{\mathbb{P}}_s, q)$ represents the optimal transport distance between the source distribution $\hat{\mathbb{P}}_s$ and the weighted distribution $q$, measured along the graph structure. This term explicitly integrates topology information to enforce minimal changes in sample densities along the manifold. 
As detailed in Section \ref{sec:optimization}, this term transfers the optimization of sample densities from Euclidean space to geometric Wasserstein space. 
Here, densities are constrained to change exclusively along the graph structure. 
This enforcement encourages \emph{local smoothness} of sample densities relative to the manifold, which helps to mitigate against potential noisy samples and edges.

\begin{algorithm}[t]
	\caption{Topology-Aware Dynamic Reweighting (TAR) Scheme} 
	\label{algo:tar}
	\begin{algorithmic}
	\STATE {\bfseries Input:} Labeled training nodes $D=\{(x_i,y_i)\}_{i=1}^N$, learning rate $\gamma$, gradient flow iterations $T_{\text{in}}$, entropy term $\beta$, graph structure $G_0=(V,E,W)$.
	\STATE {\bfseries Initialization}: Sample probability densities initialized as $(1/N, \dots, 1/N)^T$. Model parameters initialized as $\theta^{(0)}$.
	\vskip 0.05in
	\FOR{ $i=0$ {\bfseries to} $\text{Epochs}$}
		\STATE 1. Simulate gradient flow for $T_{\text{in}}$ time steps according to Equation \ref{equ:gradient} and \ref{equ:update-q} to learn an approximate worst-case probability weight $q^{T_{\text{in}}}$.
		\STATE 2. $\theta^{(i+1)} \leftarrow \theta^{(i)} - \gamma \nabla_\theta(\sum_i q^{T_{\text{in}}}_i\ell(f_\theta(x_i), y_i))$ 
	\ENDFOR	
	\end{algorithmic}
\end{algorithm}

\subsection{Optimization}
\label{sec:optimization}
The main challenge of Problem~\ref{equ:obj} lies in the computation of discrete geometric Wasserstein distance $\mathcal{GW}_{G_0}^2(\hat{\mathbb P}_s, q)$, which itself involves an complicated optimization problem and does not have an analytical form.
In this section, following Chow et al.~\cite{chow2017entropy} and Liu et al.~\citep{DBLP:conf/nips/LiuW0022}, we propose to leverage Wasserstein gradient flow to approximately solve the inner maximization problem.
The whole algorithm involves a minimax optimization, where we iteratively perform gradient ascents (on $q$) for the inner maximization and descents (on $\theta$) for the outer minimization.
The pseudo-code of our algorithm is shown in Algorithm~\ref{algo:tar}.

\noindent\textbf{Inner maximization problem.}\quad 
For easy notion, we define 
\begin{equation*}
	\mathcal L(\theta, q) := \sum_{i=1}^N q_i\ell(f_\theta(x_i),y_i) - \beta\cdot\sum_{i=1}^N q_i\log q_i.
\end{equation*}
Generally, the goal of the inner maximization problem in Equation~\ref{equ:obj} is to maximize $\mathcal L(\theta,q)$ and to minimize the topology penalty $\mathcal{GW}_{G_0}^2(\hat{\mathbb P}_s, q)$ w.r.t. sample densities $q$.
Instead of directly computing the topology penalty, we solve the inner maximization via gradient ascents on $q$ in the geometric Wasserstein space $(\mathcal P_o(G_0), \mathcal{GW}_{G_0})$, where the topology penalty $\mathcal{GW}_{G_0}^2(\hat{\mathbb P}_s, q)$ is approximated by the length of the gradient flow trajectory in the metric space.

As stated in Definition~\ref{def:gw-distance}, the continuous gradient flow is denoted by $q:[0,1]\rightarrow \mathcal P_o(G_0)$, and $q(t)$ represents the sample density at time $t\in [0,1]$.
In order to derive empirical optimization approaches, we introduce the \emph{time-discretized} gradient flow, denoted by $q^\tau: [0,T]\rightarrow \mathcal P_o(G_0)$, and the superscript $\tau$ is the value of \emph{time step} (here we introduce this superscript because different time steps refer to different time-discretized gradient flow function).
For the approximate optimization, similar with Liu et al.~\citep{DBLP:conf/nips/LiuW0022}, we leverage this time-discretized gradient flow (with time step $\tau$) of $-\mathcal L(\theta,q)$ in the geometric Wasserstein space $(\mathcal P_o(G_0), \mathcal{GW}_{G_0})$ as:
\begin{equation}
\label{equ:discrete-flow}
	q^\tau(t+\tau) \leftarrow \arg\max_{q\in\mathcal P_o(G_0)} \mathcal L(\theta,q)- \frac{1}{2\tau}\cdot \mathcal{GW}_{G_0}^2(q^\tau(t), q),
\end{equation}
which aims to obtain the ``local'' maximum of $\mathcal L(\theta,q)$ around $q^\tau(t)$ at time $t$ and meanwhile restricts the topology distance $\mathcal{GW}_{G_0}^2(q^\tau(t), q)$.
We derive the analytical form of Equation~\ref{equ:discrete-flow} as $\tau \rightarrow 0$.
For the ease of notion, the sample density of the $i$-th node at time $t$, originally denoted by $q^\tau_i(t)$, is abbreviated as $q_i(t)$, and then Equation~\ref{equ:discrete-flow} becomes:
\begin{equation}
\label{equ:gradient}
\begin{aligned}	
	\frac{dq_i(t)}{dt} &= \sum_{j: (i,j)\in E}w_{ij}v_{ij} \bigg(\mathbb{I}(v_{ij}>0)q_j + \mathbb{I}(v_{ij}\leq 0)q_i\bigg)\\
	v_{ij} &= \ell_i-\ell_j+\beta(\log q_j - \log q_i), \quad\text{for }(i,j)\in E
\end{aligned}
\end{equation}
where $E$ is the edge set of graph $G_0$, $w_{ij}$ is the edge weight between node $i$ and $j$, $\mathbb{I}(\cdot)$ is the indicator function, and $\ell_i$ represents the prediction error on the $i$-th node.
Intuitively, $v_{ij}$ can be viewed as the transferring velocity of the sample density from node $j$ to node $i$.

\begin{figure*}
	\centering\includegraphics[width=0.9\textwidth]{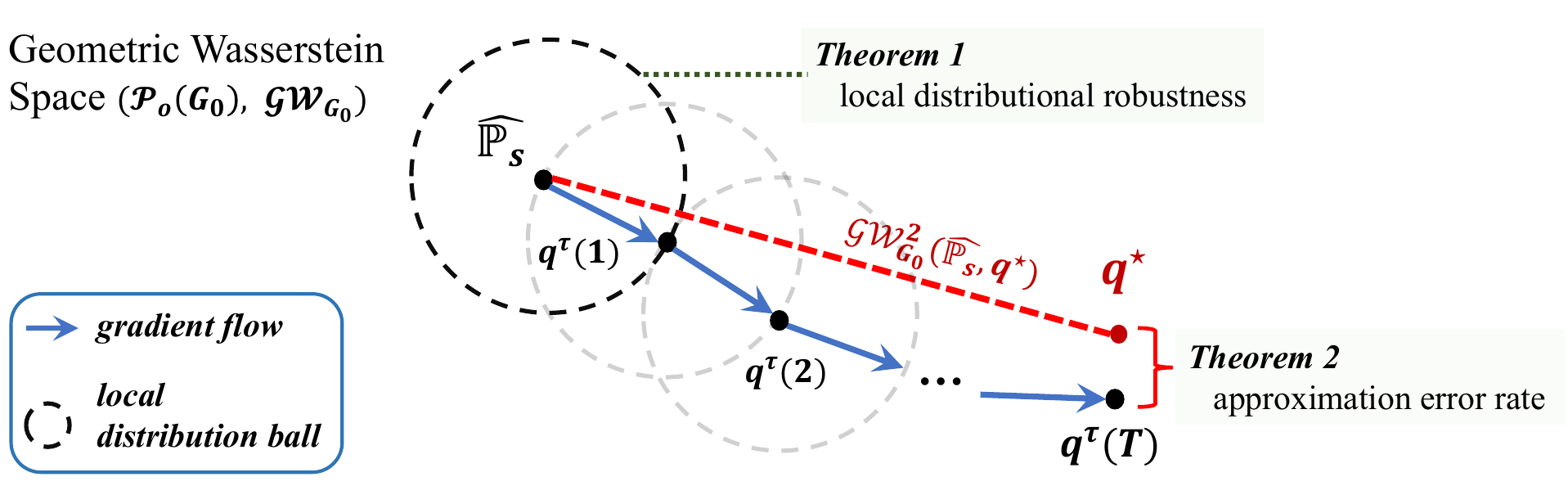}
	\caption{Illustration of the gradient flow in the geometric Wasserstein space $(\mathcal P_o(G_0), \mathcal{GW}_{G_0})$, where each point denotes a probability distribution in $\mathcal P_o(G_0)$, and the distance is measure by the discrete geometric Wasserstein distance. The black circle denotes the local distribution set around a distribution, and the blue arrow represents the one-step gradient flow. $q^\tau(T)$ denotes the approximated inner maximizer obtained by our algorithm, and $q^\star$ denotes the ground-truth inner maximizer (defined in Theorem \ref{theorem:error}). In Theorem~\ref{theorem:dro}, we demonstrate that the one-step gradient flow is equivalent to distributionally robust optimization around a local uncertainty set, and in Theorem~\ref{theorem:error}, we characterize the approximation error rate between $q^\tau(T)$ and $q^\star$.}
	\label{fig:gf}
\end{figure*}

Let $\lambda=\frac{1}{2\tau}$, Equation~\ref{equ:discrete-flow} exactly aligns with the goal of our inner maximization problem in Problem~\ref{equ:obj}.
Specifically, the original topology penalty calculates the distance $\mathcal{GW}_{G_0}^2(\hat{\mathbb{P}}_s, q^\tau(t))$ between $\hat{\mathbb P}_s$ and $q^\tau(t)$, and our gradient flow approximates it via $\sum_{i=1}^t \mathcal{GW}^2_{G_0}(q^{\tau}(i-1), q^\tau(i))$ (see blue curves in Figure~\ref{fig:gf}).
In Theorem~\ref{theorem:error}, we characterize the error rate of this approximation.

\begin{remark}
	Here we make some remarks on Equation~\ref{equ:gradient}:\\ 
	(a) The gradient of the $i$-th node's probability density depends on its neighbors in graph $G_0$. 
	This corresponds with our motivation that the reweighting scheme should incorporate topology information. 
	Furthermore, since the transfer is between neighbors, the probability density $p$ remains \emph{locally smooth} w.r.t. the graph structure (or manifold), which avoids overemphasis on some noisy samples.\\
	(b) Combined with our topology penalty, the entropy penalty acts as a \emph{non-linear graph Laplacian operator} to further the smoothness of probability densities along the manifold.\\
	(c) The gradient flow in Equation~\ref{equ:gradient} is implemented by message propagation, which \emph{scales linearly with sample size} and enjoys parallelization by GPU.\\
        (d) Due to the random sampling of labeled nodes during training for node classification tasks, it means that for certain nodes we cannot compute the loss, which disrupts the connectivity and hinders the calculation of this Equation, we intuitively set the loss for these unlabeled nodes to the mean loss of the labeled nodes, and this approach has proven to be adequate. For other potential solutions, please refer to the Appendix~\ref{sec:reconnect}.
\end{remark}

Based on Equation~\ref{equ:gradient}, we can solve the inner maximization problem via gradient ascent as:
\begin{align}
\label{equ:update-q}
	& q_i(0) \leftarrow 1/N, \\
	& q_i(t+1) \leftarrow q_i(t) + \tau\cdot dq_i(t)/ dt, \quad\text{for }i\in [N].
\end{align}

In addition, we demonstrate the equivalence between Equation~\ref{equ:discrete-flow} and distributional robustness in Theorem~\ref{theorem:dro}, justifying how our proposed TAR method can provide robustness against distribution shifts.
And in Theorem~\ref{theorem:error}, we characterize the error rate of our approximation as $e^{-CT_{\text{in}}}$, which allows a relatively accurate approximation with finite $T_{\text{in}}$ steps.

\noindent\textbf{Outer minimization problem.}\quad 
For the outer minimization problem, we perform gradient descent on model parameters $\theta$.
According to the overall objective in Equation~\ref{equ:obj}, the loss function is simply a weighted average:
\begin{equation}
\label{equ:theta}
	\theta^{(t+1)} \leftarrow \theta^{(t)} - \gamma\cdot \nabla_\theta \bigg(\sum_{i=1}^N q_i(T_{\text{in}})\cdot\ell(f_\theta(x_i), y_i)\bigg),
\end{equation}
where $\gamma$ is the learning rate, and $q_i(T_{\text{in}})$ denotes the probability density of the $i$-th node (after $T_{\text{in}}$ steps gradient flow).

\subsection{Theoretical Analysis}
\label{sec:theoretical}
In this section, we investigate in-depth our proposed optimization algorithm.
As illustrated in Figure~\ref{fig:gf}, we first prove that each step of the gradient flow exactly finds the worst-case distribution within a local uncertainty set (see black circle in Figure~\ref{fig:gf}).

\begin{theorem}[Distributional robustness]
\label{theorem:dro}
	For any $\gamma > 0, t>0$ and given $\theta$, denote the solution of Equation \ref{equ:discrete-flow} as $q^\star=\arg\max_{q\in\mathscr{P}_o(G_0)}\mathcal{L}(\theta,q)-\gamma\mathcal{GW}_{G_0}^2(p,q)$.
	Let $\epsilon=\mathcal{GW}^2_{G_0}(p,q^\star)$, we have
	\begin{equation}
		\underbrace{\max_{q\in\mathscr{P}_o(G_0)}\mathcal{L}(\theta,q)-\gamma\mathcal{GW}_{G_0}^2(p,q)}_{\text{one-step gradient flow at time $t$}} = \underbrace{\max_{q:\mathcal{GW}^2_{G_0}(p,q)\leq \epsilon}\mathcal{L}(\theta,q).}_{\text{the worst-case distribution within a local distribution set}}
	\end{equation}	
	The proof can be found in Appendix~\ref{sec:proof}.
\end{theorem}
Theorem~\ref{theorem:dro} shows that, for the inner maximization, our proposed gradient flow is equivalent to finding the worst-case distribution within a small distribution set.
Therefore, the weighted average loss function in Equation~\ref{equ:theta} captures the worst-case distribution that may occur in testing, which shares the similar idea with distributionally robust optimization~\cite{DBLP:journals/corr/abs-1810-08750, DBLP:journals/jap/BlanchetKM19, DBLP:conf/nips/LiuW0022}.
This demonstrates the strength of our proposed TAR framework in dealing with potential distribution shifts.

Then based on the results in~\citep[Theorem 5]{chow2017entropy} and~\citep[Theorem 3.2]{DBLP:conf/nips/LiuW0022}, we move on to analyze the error rate of our approximation in Theorem~\ref{theorem:error}.

\begin{theorem}[Approximation error rate]
\label{theorem:error}
Given the GNN parameter $\theta$, denote the approximate sample densities in Equation \ref{equ:gradient} after $T_{\text{in}}$ steps of gradient flow as $q(T_{\text{in}})$, and $\epsilon=\mathcal{GW}_{G_0}^2(\hat{\mathbb P}_{s}, q(T_{\text{in}}))$ is the geometric Wasserstein distance from the original source distribution.
Denote the ground-truth worst-case distribution with the same distance $\epsilon$ as:
\begin{equation*}
	q^\star=\arg\max\limits_{q:\mathcal{GW}_{G_0}^2(\hat{\mathbb P}_{s},q)\leq \epsilon}	\mathcal L(\theta,q),
\end{equation*}
Then we have:
	\begin{equation}
	\label{equ:error-rate}
		\frac{\mathcal{L}(\theta, q(T_{\text{in}})) - \mathcal{L}(\theta,\hat{\mathbb P}_{s})}{\mathcal{L}(\theta,q^\star)-\mathcal{L}(\theta,\hat{\mathbb P}_{s})} > 1-e^{-CT_{\text{in}}},
	\end{equation}	
where $C>0$ is a constant and its value depends on the loss function $\ell$, hyper-parameter $\beta$, and sample size $N$.
The proof can be found in Appendix~\ref{sec:proof}.
\end{theorem}

\begin{remark}
We make some remarks here:\\
(1) Since the goal of our reweighting is to maximize $\mathcal{L}(\theta,q)$ w.r.t. $q$, we utilize the increase of $\mathcal L$ to characterize how ``approximate'' is our optimization.  
	In Equation~\ref{equ:error-rate}, the denominator of the left-hand side represents the maximal increase, and the numerator is the increase attained through our approximation.
	As the ratio approaches 1.0, our approximation becomes increasingly precise.\\
(2) Our theoretical results show that the error rate is $e^{-CT_{\text{in}}}$, which shrinks fast as the number of time step $T_{\text{in}}$ increases.
	This further demonstrates that our optimization is able to find good approximations in finite (usually small) number of gradient flow steps.
\end{remark}

%% file: data/4exp.tex
\section{Experiment}\label{sec:exp}


We conduct experiments on four OOD node classification datasets under both concept shift and covariate shift to validate the effectiveness of our proposed method. 
Additionally, we evaluate our method on three long-tailed node classification datasets to assess its effectiveness in addressing class imbalance tasks.
\subsection{Datasets and Baselines}
\paragraph{Datasets.} \textbf{(1)} For OOD datasets, We use four node classification datasets under both concept shift and covariate shift: WebKB~\cite{pei2020geom}, CBAS~\cite{ying2019gnnexplainer}, Twitch~\cite{rozemberczki2020characteristic}, and Cora~\cite{bojchevski2017deep}. We followed the GOOD benchmark~\cite{gui2022good} for data splitting. Specifically:
\begin{itemize}
    \item \textbf{WebKB} is a five-class dataset for classifying web pages into different categories, constructed to exhibit distribution shift through different university domains.
    \item \textbf{CBAS} is a synthetic four-class dataset that induces distribution shift via node colors.
    \item \textbf{Twitch} is a binary classification gamer network dataset, where each node represents a gamer, and distribution shift is introduced through gamer language.
    \item \textbf{Cora} is a seventy-class citation dataset, with each node representing a paper, and distribution shift is introduced through the selected word count of each paper.
\end{itemize}

\textbf{(2)} For class-imbalanced datasets, we validate the performance of TAR in a class-imbalanced setting on three benchmark datasets (Cora, CiteSeer, PubMed). Following the partitioning approach of GraphENS \cite{park2021graphens}, we construct a long-tail citation network to validate TAR under a high imbalance ratio, which represents the ratio between the most frequent class and the least frequent class. In our experiments, we set this imbalance ratio to 100.

\paragraph{Baselines.}\textbf{(1)} For OOD experiments, we use ERM and general domain generalization baselines, including IRM~\cite{arjovsky2019invariant}, VREx\cite{krueger2021out}, Group DRO\cite{sagawa2019distributionally}, DANN~\cite{ganin2016domain}, and Deep Coral\cite{sun2016deep}. Additionally, we include graph-specific domain generalization baselines such as EERM~\cite{EERM} and SR-GCN~\cite{SRGNN}. In these domain generalization methods, except for DANN, IRM, and EERM, all others require domain labels to help address distribution shifts, and our proposed TAR does not require domain labels either. \textbf{(2)} For class-imbalanced experiments, we compare our TAR with several reweighting-based approaches for handling class imbalance, using both GAT \cite{gat} and oversampling-based GraphENS \cite{park2021graphens} as backbone models. Specifically, we compare with Re-Weight \cite{reweight}, which scales class weights proportional to the number of class samples; Class-Balanced Loss (CB Loss) \cite{cbloss}, a generic method that modifies the loss function to address imbalance issues; and TAM \cite{tam}, a state-of-the-art node-wise logit adjustment method for handling node class,  which aims to decrease the false positive cases considering the topological structure of graphs. To ensure fairness in comparisons, we use the same backbone model parameters, layer configurations, and random seeds across all experiments. For more training details, please refer to Appendix~\ref{sec:ci_setting}.

\paragraph{Evaluation Metrics.}\textbf{(1)} For OOD experiments, we use Accuracy (Acc) as the evaluation metric for all tasks except for the Twitch dataset, where we employ ROC-AUC as the evaluation metric. \textbf{(2)} For class-imbalanced experiments, we use the Accuracy, average balanced precision (bAcc), and F1 score. For more detailed descriptions related to the evaluation metrics, please refer to the Appendix~\ref{sec:metrics}.

\input{./data/table/comparison}

\input{./data/image/step}

\input{./data/table/imbalance}

\subsection{Performance Comparison on GOOD Benchmark}
Table~\ref{tab:comparison} summarizes the results of our method and other baselines on four datasets under both covariate shift and concept shift.
Our proposed TAR method outperforms all baselines in 6 out of 8 standard settings and achieves the second-best performance in the remaining 2 settings.

We have the following observations:
Among the domain generalization baselines, EERM demonstrates excellent performance on the WebKB dataset under covariate shift while it fails under concept shift. Besides, its performance is not stable, as it performs worse than ERM on the CBAS dataset and encounters OOM (Out of Memory) issues in our experimental setup. DANN achieves the best performance under covariate shift on the Twitch dataset. However, under concept shift on both the Twitch and WebKB datasets, its performance is merely on par with ERM. We speculate that this is because DANN is designed to extract transferable features from different domains but lacks mechanisms specifically addressing concept shift. None of the baselines surpass ERM across all datasets under both concept shift and covariate shift, this reveals that simply applying generic Out-of-Distribution (OOD) generalization methods to graphs is not effective in solving OOD node classification tasks, and current graph OOD generalization methods fail to deal with concept shift and covariate shift simultaneously.

As shown in Table~\ref{tab:comparison}, compared with ERM, our proposed method TAR obtains consistent improvements across all datasets under both concept shift and covariate shift and achieves the best results in 6 out of 8 dataset settings, demonstrating its effectiveness in dealing with distribution shifts. For CBAS under covariate shift, TAR achieves a performance improvement of 1.15\% compared to the best baseline Group DRO and IRM. Specifically, for CBAS under concept shift, we note that all baselines underperform ERM, while TAR obtains a 0.71\% improvement, which illustrates that our method can overcome the shortcomings of both general domain generalization methods and existing graph domain generalization methods when tailored for OOD node classification tasks. Moreover, TAR alleviates distribution shifts without requiring domain labels, making it more feasible for real-world scenarios compared to methods like Group DRO, VREx, IRM, SRGNN, and Deep Coral, which require domain labels to address distribution shifts.

\subsection{Hyper-Parameter Analysis}
We analyze the impact of $T_{\text{in}}$ (the number of iterations for adjusting sample weights) and $\beta$ (the smoothness of sample weights within neighborhoods) on the performance of TAR (exclusively under concept shift). Specifically, $T_{\text{in}}$ ranged from \{1, 3, 5, 10, 30, 100\} and $\beta$ ranged from \{1, 0.1, 0.01, 0.001, 0\}.

For $T_{\text{in}}$, since the sample weights must transfer along the edges of the graph, the value of $T_{\text{in}}$ determines how many iterations the sample weights can transfer. A larger $T_{\text{in}}$ means that the weights can propagate over a wider range, while a smaller $T_{\text{in}}$ means that the weights can only transfer to a few neighboring hops. As shown in Figure~\ref{fig:step}, the performance saturates at $T_{\text{in}} = 10$ for the WebKB and CBAS datasets, whereas for the Twitch dataset, the performance saturates at $T_{\text{in}} = 30$. We speculate that this is because Twitch is a larger graph compared to WebKB and CBAS (Twitch has 34,120 nodes, while WebKB and CBAS have 617 and 700 nodes, respectively), thus requiring more iterations for the sample weights to transfer effectively.

For $\beta$, it controls the smoothness of sample weights within the neighborhood. A larger $\beta$ means that adjacent samples will have more similar weights, while a smaller $\beta$ means that the weights of adjacent samples can differ more significantly, which could lead to some noisy samples attracting too much attention during training. As shown in Figure~\ref{fig:step}, a $\beta$ value of 0.01 yields better results.



\subsection{Performance Comparison on Class-Imbalanced Setting}
We also conduct experiments on class-imbalanced node classification tasks to validate the effectiveness of TAR in addressing class imbalance tasks. In Table~\ref{tab:imbalance}, we report the average test accuracy (Acc), average balanced precision (bAcc), and F1 score in terms of their standard deviation for the baseline method and TAR on the three long-tailed partitioned citation networks \cite{park2021graphens}. For the base ERM method \cite{gat}, we compare it with different reweighting-based methods. The experimental results show that our method is either the best or second best, and the bAcc on PubMed dataset can achieve approximately 10\% improvement. Experimental integration with GraphENS \cite{park2021graphens}, a state-of-the-art (SOTA) oversampling-based imbalance method, also achieves competitive results on all the datasets. In addition, our proposed method can be integrated into any of the class-imbalanced handling methods of SOTA to achieve more competitive results.

%% file: data/table/comparison.tex
\begin{table}[!t]
\caption{The performance on four OOD benchmark datasets. We report the average test accuracy and standard deviations over 10 runs. The best results are shown in \textbf{bold}, and the second best results are shown in \underline{underline}. OOM denotes
out of memory.}
\vspace{0.1in}
    \centering
    \resizebox{\textwidth}{!}{
    \begin{tabular}{c|cccccccc}
    \toprule
        Dataset & \multicolumn{2}{c}{WebKB} & \multicolumn{2}{c}{CBAS} & \multicolumn{2}{c}{Twitch} & \multicolumn{2}{c}{Cora} \\ \midrule
        Shift & concept & covariate & concept & covariate & concept & covariate & concept & covariate \\ \midrule
        ERM & 26.97$\pm$1.49 & 14.13$\pm$2.92 & \underline{82.86$\pm$1.28} & 78.43$\pm$1.00 & 47.87$\pm$0.65 & 48.55$\pm$0.91 & 64.41$\pm$0.35 & 64.56$\pm$0.35 \\ 
        IRM & 27.62$\pm$1.50 & 17.14$\pm$7.13 & 82.72$\pm$0.95 & \underline{78.71$\pm$1.86} & 48.16$\pm$0.63 & 48.16$\pm$0.59 & 64.45$\pm$0.34 & 64.51$\pm$0.33 \\ 
        VREx & 27.07$\pm$1.44 & 16.03$\pm$6.45 & 82.29$\pm$1.46 & 78.43$\pm$1.86 & \underline{48.53$\pm$0.86} & 47.83$\pm$0.48 & 64.48$\pm$0.20 & 64.33$\pm$0.26 \\ 
        Group DRO & 26.88$\pm$1.36 & 14.44$\pm$4.15 & 82.79$\pm$1.34 & \underline{78.71$\pm$1.62} & 47.74$\pm$0.68 & 48.95$\pm$1.17 & 64.45$\pm$0.42 & \underline{64.62$\pm$0.33} \\ 
        DANN & 26.97$\pm$1.93 & 15.87$\pm$3.94 & 81.93$\pm$1.28 & 78.14$\pm$2.40 & 47.87$\pm$0.62 & \textbf{51.08$\pm$3.07} & 64.44$\pm$0.38 & 64.59$\pm$0.34 \\ 
        Deep Coral & 26.88$\pm$1.54 & 13.97$\pm$3.66 & 82.43$\pm$1.07 & 78.14$\pm$2.93 & 47.86$\pm$0.61 & 48.13$\pm$0.62 & 64.51$\pm$0.32 & 64.57$\pm$0.34 \\ \midrule
        EERM & 26.88$\pm$1.64 & \textbf{26.59$\pm$10.17} & 64.14$\pm$2.41 & 61.00$\pm$12.02 & OOM & OOM & OOM & OOM \\ 
        SRGNN & \underline{27.80$\pm$1.88} & 13.89$\pm$2.28 & 81.57$\pm$0.70 & 73.86$\pm$2.93 & 47.94$\pm$0.67 & 48.47$\pm$0.70 & \underline{64.76$\pm$0.24} & 64.27$\pm$0.33 \\ \midrule
        TAR (ours) & \textbf{27.98$\pm$1.02} & \underline{18.57$\pm$3.30} & \textbf{83.57$\pm$1.57} & \textbf{79.86$\pm$2.07} & \textbf{49.32$\pm$0.63} & \underline{49.20$\pm$1.39} & \textbf{64.79$\pm$0.26} & \textbf{64.78$\pm$0.19} \\ \bottomrule
    \end{tabular}
    }
    \label{tab:comparison}
\end{table}

%% file: data/image/step.tex

\begin{figure}[t]
  \vspace{-1.5cm}
  \subfloat[$T_\text{in}$ on CBAS]{\includegraphics[width=0.33\textwidth, height=0.24\textwidth]{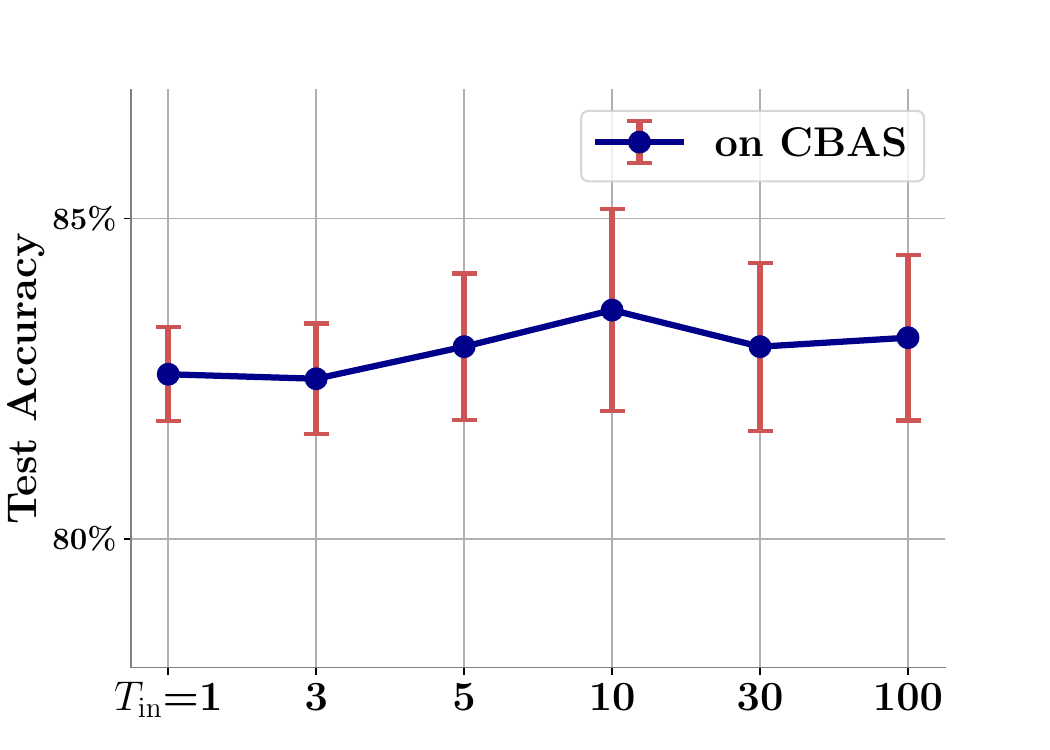}\label{fig:overall_subfig1}}
  \subfloat[$T_{\text{in}}$ on Twitch]{\includegraphics[width=0.33\textwidth, height=0.24\textwidth]{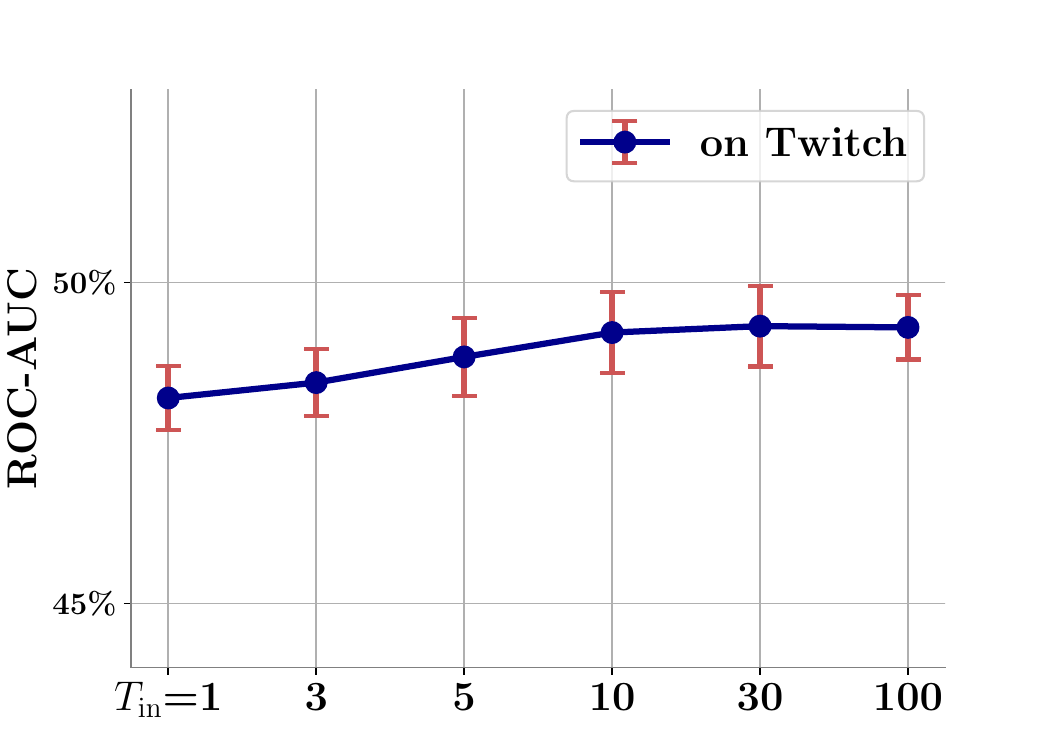}\label{fig:overall_subfig3}}
  \subfloat[$T_{\text{in}}$ on WebKB]{\includegraphics[width=0.33\textwidth, height=0.24\textwidth]{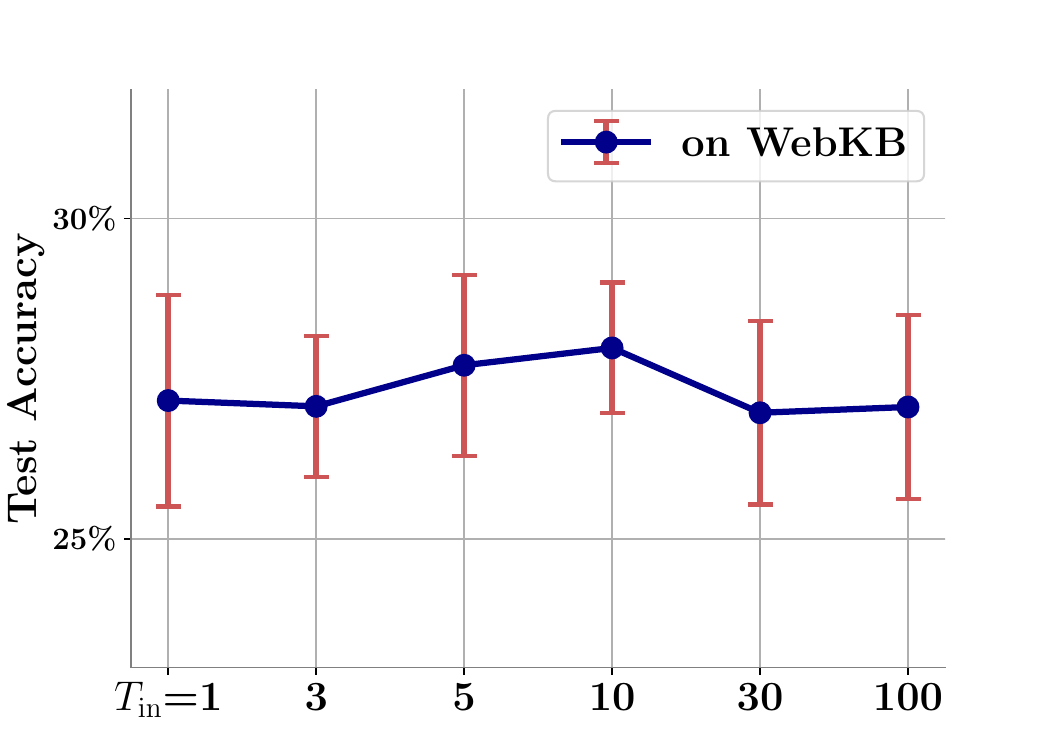}\label{overall_fig:subfig4}}
  \hfill
  \vspace{-0.15in}
  \subfloat[$\beta$ on CBAS]{\includegraphics[width=0.33\textwidth, height=0.24\textwidth]{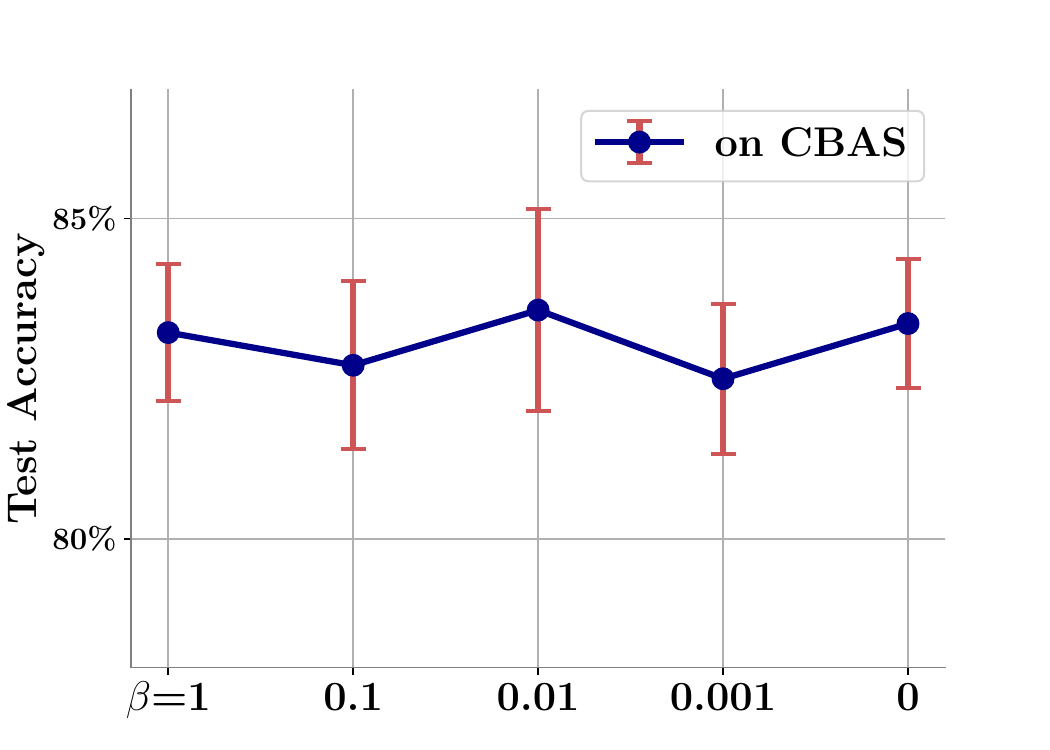}\label{fig:overall_subfig5}}
  \subfloat[$\beta$ on Twitch]{\includegraphics[width=0.33\textwidth, height=0.24\textwidth]{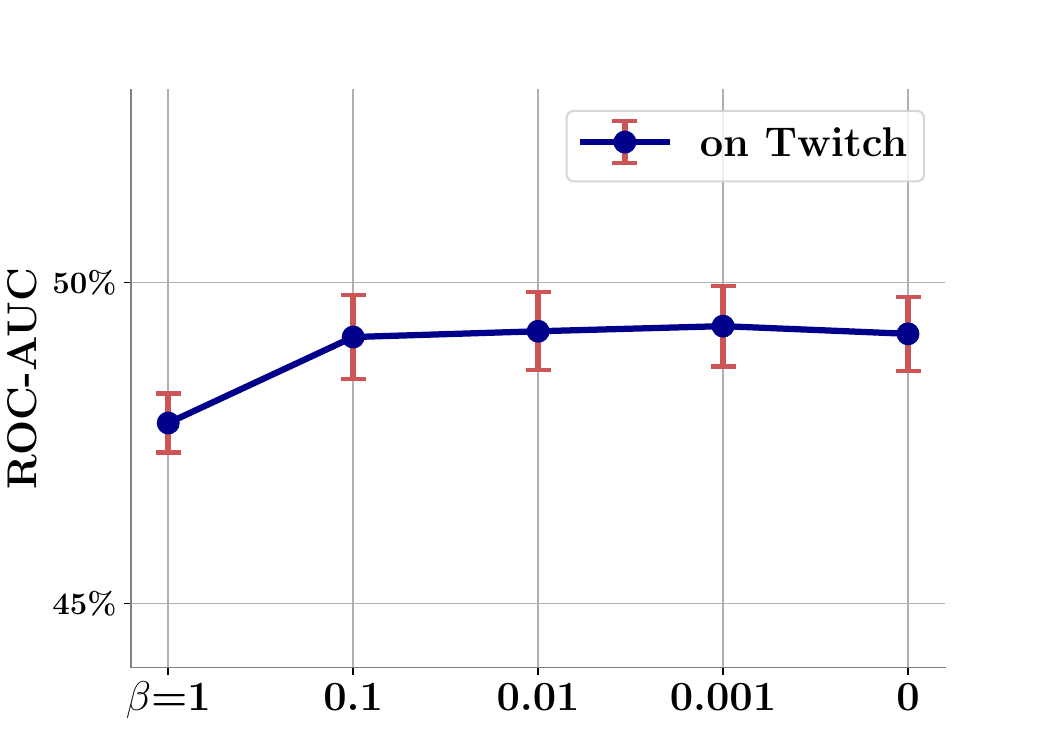}\label{fig:overall_subfig6}}
 \subfloat[$\beta$ on WebKB]{\includegraphics[width=0.33\textwidth, height=0.24\textwidth]{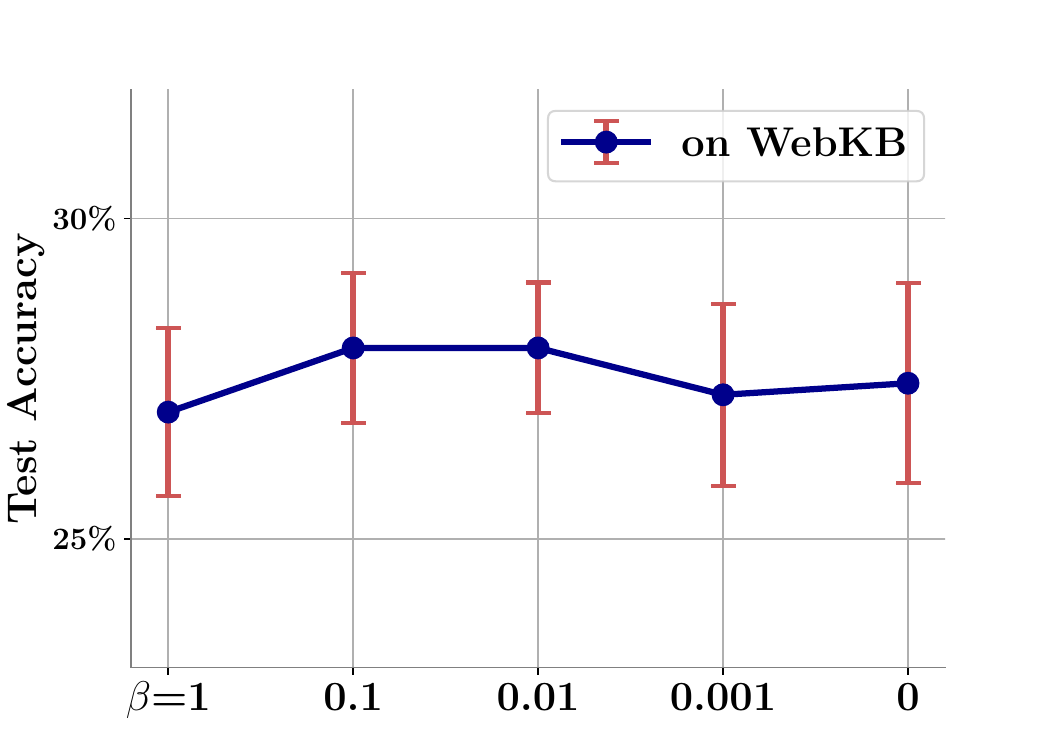}\label{fig:overall_subfig7}}

  \caption{The effects of hyper-parameters $T_{\text{in}}$ (the number of gradient flow) and $\beta$ (the coefficient of the entropy penalty) of our proposed TAR algorithm.}
 \label{fig:step}  
\end{figure}

%% file: data/table/imbalance.tex
\begin{table}[t]
\caption{Comparison of TAR with other baselines in class-imbalanced settings. We report the averaged accuracy (Acc), balanced accuracy (bAcc), and F1-score (F1) with the standard deviations for 5 repetitions on three datasets. The best results are shown in \textbf{bold}, and the second best results are shown in \underline{underline}.}
\vspace{0.1in}
\resizebox{\textwidth}{!}{
\begin{tabular}{lccc ccc ccc}
\toprule
Dataset & \multicolumn{3}{c}{Cora-LT} & \multicolumn{3}{c}{CiteSeer-LT} & \multicolumn{3}{c}{PubMed-LT} \\
 Metric        & Acc & bAcc & F1 & Acc & bAcc. & F1 & Acc & bAcc & F1 \\
        \midrule
ERM & 73.04$\pm$0.15 & 63.83$\pm$0.32 & 63.67$\pm$0.60& 54.38$\pm$0.36 & 47.83$\pm$0.36   & 43.58$\pm$0.57 & 70.80$\pm$0.39 & 57.77$\pm$0.32 & 52.58$\pm$0.33 \\
w/ Re-Weight &73.40$\pm$0.30 & 64.51$\pm$0.43 & 64.85$\pm$0.38 & 53.84$\pm$0.44  & 47.31$\pm$0.41   & 42.79$\pm$0.66   & \underline{71.00$\pm$0.27}  & 57.83$\pm$0.24 & 52.37$\pm$0.30 \\
w/ CB Loss  & 73.10$\pm$0.25 & 63.59$\pm$0.35 & 63.55$\pm$0.96 & 54.93$\pm$0.15   & 48.39$\pm$0.21   & 44.40$\pm$0.37   & 70.92$\pm$0.24 & 57.76$\pm$0.21 & 52.32$\pm$0.26 \\
w/ TAM  & \textbf{74.62$\pm$0.26} & \underline{65.03$\pm$0.69}  & \underline{65.44$\pm$0.77}  & \underline{56.82$\pm$0.23} & \underline{49.97$\pm$0.20} & \underline{44.94$\pm$0.29} & 70.80$\pm$0.48 & \underline{59.13$\pm$0.52} & \underline{56.34$\pm$0.74}   \\
w/ TAR (Ours)  & \underline{74.17$\pm$0.38} & \textbf{66.10$\pm$0.76} & \textbf{66.29$\pm$0.58} & \textbf{57.08$\pm$0.53}   & \textbf{50.41$\pm$0.56}   & \textbf{47.01$\pm$0.77}  & \textbf{75.40$\pm$0.43} & \textbf{67.33$\pm$0.59} & \textbf{68.23$\pm$0.75} \\
\midrule
GraphENS& 77.64$\pm$0.15 & 72.23$\pm$0.27 & 72.12$\pm$0.33 & 62.30$\pm$0.35 & 56.07$\pm$0.30 & 54.33$\pm$0.43 & 76.56$\pm$0.44 & 71.05$\pm$1.04 & 71.98$\pm$0.96 \\
w/ Re-Weight& 77.86$\pm$0.16 & 72.47$\pm$0.34 & 72.77$\pm$0.41 & 62.50$\pm$0.15 & 56.28$\pm$0.18 & 54.76$\pm$0.23 & 77.32$\pm$0.72 & \underline{72.08$\pm$1.41} & \underline{73.09$\pm$1.36}   \\
w/ CB Loss& 77.68$\pm$0.40 & 72.79$\pm$0.62 & 72.91$\pm$0.72 & 63.38$\pm$0.84 & 56.93$\pm$0.73 & 55.29$\pm$0.84 & 77.18$\pm$0.51 & 70.83$\pm$1.07 & 71.98$\pm$1.13 \\
w/ TAM& \textbf{78.86$\pm$0.20} & \textbf{73.28$\pm$0.38} & \textbf{73.57$\pm$0.41} & \underline{63.80$\pm$0.63} & \underline{57.31$\pm$0.55} & \underline{55.51$\pm$0.50} & \underline{77.90$\pm$0.21}&71.70$\pm$0.29 & 73.03$\pm$0.27 \\
w/ TAR (Ours)& \underline{78.34$\pm$0.23} & \underline{73.16$\pm$0.25} & \textbf{73.57$\pm$0.39} & \textbf{64.96$\pm$0.39} & \textbf{58.39$\pm$0.41} & \textbf{56.64$\pm$0.49} & \textbf{78.56$\pm$0.47}&\textbf{73.25$\pm$0.82} & \textbf{74.40$\pm$0.81} \\ 
\bottomrule
\end{tabular}
}
\label{tab:imbalance}
\end{table}

%% file: data/5con.tex
\section{Conclusion}
Through this work, we innovatively propose the Topology-Aware Dynamic Reweighting (TAR) framework to address the distribution shift problem in node classification tasks. 
TAR utilizes a minimax approach to enhance the generalization ability of GNN models, incorporating topological structure information through gradient flows in the geometric Wasserstein space. We further conduct theoretical analysis to reveal the ability of TAR to enhance the distributional robustness of the GNN model. 
Experimental results confirm the effectiveness on real-world datasets of node classification. 
Our TAR opens a new direction for addressing the distribution shift problem for node classification tasks.

%% file: data/6appendix.tex
\appendix

\section{Related Works}

\paragraph{Out of Distribution Generalization.}
Out-of-Distribution (OOD) generalization aims to address the challenge of ensuring model robustness and generalization when faced with data that differ from the training distribution. Numerous studies have been dedicated to addressing the problem of OOD generalization, leading to the development of various methods for tackling OOD issues\cite{liu2021towards}. By accurately identifying the causal relationships between features and their corresponding labels, causal learning methods are expected to perform well even when the data distribution changes, as the underlying causal structure is often assumed to remain invariant across different environments or domains. Shifting the focus from strict causality to invariance, invariant learning aims to develop a representation or model that remains consistent across various environments. Invariant Risk Minimization (IRM)\cite{arjovsky2019invariant} and Variance Risk Extrapolation (VREx)~\cite{krueger2021out} are two prominent methods specifically designed to address these challenges. IRM focuses on learning invariant features by ensuring that the optimal classifier remains the same across different environments, whereas VREx aims to minimize the variance of risks across environments, ensuring stable performance under distributional shifts. Another line of research focused on addressing OOD generalization problems involves distributionally robust optimization methods. These model-agnostic techniques come with strong theoretical guarantees and achieve OOD generalization by incorporating distributional robustness into the training process. This ensures that the model's performance remains stable across different data distributions. KLDRO~\cite{DBLP:journals/corr/abs-1810-08750} minimizes the KL divergence between training and potential test distributions. WDRO~\cite{wdro1,wdro2} leverages the Wasserstein distance to ensure robustness to distributional changes. Group DRO~\cite{sagawa2019distributionally} aims to provide consistent performance across different subgroups by minimizing the worst-case risk among them. While invariant learning has been extensively applied in graph tasks\cite{EERM,xia2024learning,li2022learning,sui2023unleashing,wu2022discovering}, there is relatively less application of distributionally robust optimization methods in graph tasks. Applying these methods to graphs requires addressing the unique structural properties of graphs, posing challenges that are specific to graph data.

\paragraph{Graph Invariant Learning.}
Recently, graph invariant learning has shown enormous success in addressing graph out-of-distribution problems\cite{wu2022discovering,EERM,sui2023unleashing,ood_link,li2022learning,xia2024learning}. Graph invariant learning aims to exploit the invariant relationships between graph features(which can be divided into topological structures and node features) and labels across distribution shifts, while filtering out the variant spurious correlations caused by the environment. Recently, many methods have been proposed for graph-level tasks. GIL~\cite{li2022learning} captures the invariant relationships between predictive graph structural information and labels in a mixture of latent environments. DIR~\cite{wu2022discovering} selects a subset of causal rationales and conducts data augmentation to create multiple distributions to improve generalization. MoleOOD~\cite{yang2022learning} enhances the robustness of molecule learning and infers the environment in a fully data-driven manner. AIA~\cite{sui2023unleashing} generates new environments while preserving the original stable features during the augmentation process with adversarial strategies. Compared to research on graph-level ood, little attention has been paid to learning node-level representations under distribution shifts from the invariant learning perspective~\cite{EERM,xia2024learning}. EERM~\cite{EERM} leverages multiple context explorers that are adversarially trained to maximize the variance of risks from multiple virtual environments to learn a node invariant predictor. CIT~\cite{xia2024learning} generates nodes across different clusters, significantly enhances the diversity of the nodes and helps GNNs learn the invariant representations. However, this line of invariant learning typically focuses on specific types of invariance (e.g., subgraph invariance in graphs), which may not cover all possible shifts. Besides, due to the lack of environment information in real-world, generating new samples might introduce bias or noise.

\section{Experimental Details}\label{sec:setting}
\paragraph{OOD datasets.} In this paper, we use four OOD node classification datasets from GOOD benchmark~\cite{gui2022good}, including WebKB, CBAS, Twitch, and Cora. Statistics of each dataset are shown in Table~\ref{tab:ood_datasets}. Since the purpose of this paper is to address OOD problems, we save the model that performs best on the OOD validation set during our experiments and report its results on the OOD test set.

\paragraph{Detail setting on OOD dataset.} \label{sec:ood_setting}
For data splitting, we follow the settings of the GOOD benchmark~\cite{gui2022good}. For all baselines and our method, we conduct grid search as defined by the GOOD Benchmark and reported their best results. Note that the graph OOD algorithm EERM encounters CUDA out of memory on Twitch and Cora datasets due to its high memory requirement. For all the experiments, we use the Adam optimizer, with a weight decay of 0. We adopt the same backbone from the implementation of GOOD benchmark\footnote{\url{https://github.com/divelab/GOOD}}. The model structures and hyperparameters we used are summarized in Table~\ref{tab:ood_setting}.

\input{./data/table/ood_datasets}

\input{./data/table/ood_setting}

\input{./data/table/ci_datasets}

\input{./data/table/ci_setting}

\paragraph{Class-imbalanced datasets.} In this paper, we use three node classification datasets, including Cora, Citeseer, and PubMed. Following GraphENS~\cite{park2021graphens}, We construct a corresponding class-imbalanced train set for each dataset. The imbalance ratio, which is the ratio of the number of samples of the most frequent class $n_{major}$ and the number of samples of the least frequent class $n_{minor}$, is set to 100. The detailed label distribution of the train set for each dataset are shown in Table~\ref{tab:ci_datasets}.

\paragraph{Detail setting on class-imbalanced dataset.}\label{sec:ci_setting}
 In our experiments, we use the code implementation of GraphENS\footnote{\url{https://github.com/JoonHyung-Park/GraphENS}}, and then integrate the implementations of TAM and our method into it. For different methods, we always use the same backbone. For all the experiments, we use the Adam optimizer, with a weight decay of 0. The model structures and hyperparameters we used are summarized in Table~\ref{tab:ci_setting}.

\paragraph{Software and Hardware.} Our implementation is under the architecture of PyTorch~\cite{torch} and PyG~\cite{pyg}. All of our experiments are run on one GeForce RTX 2080Ti with 12GB. The detailed versions of some key packages are listed below:
\begin{itemize}
    \item python: 3.8
    \item pytorch: 1.10.1
    \item pyg: 2.0.3
\end{itemize}

\section{Evaluation Metrics}\label{sec:metrics}
When dealing with the class imbalance setting, we follow GraphENS and choose accuracy, balanced accuracy and F1 score as evaluation metrics. Accuracy (Acc), balanced precision (bAcc), and F1-score (F1) are explained in detail below:

\textbf{Accuracy (Acc)}: Only using accuracy can be misleading in the presence of class imbalance. For example, if 95\% of the samples belong to the majority class, a model that always predicts the majority class will have 95\% accuracy, but it won't perform well on the minority class.

\begin{equation*}
\text{Accuracy} = \frac{TP + TN}{TP + TN + FP + FN}
\end{equation*}
where TP denotes True Positives, TN debotes True Negatives, FP denotes False Positives and FN denotes False Negatives.

\textbf{Balanced Accuracy (bAcc)}:Balanced Accuracy is the average of recall obtained on each class. It is used to deal with imbalanced datasets by taking into account both sensitivity (recall for the positive class) and specificity (recall for the negative class).
\begin{equation*}
\text{Balanced Accuracy} = \frac{1}{2} \left( \frac{TP}{TP + FN} + \frac{TN}{TN + FP} \right)
\end{equation*}

\textbf{F1 Score}: The F1 Score is the harmonic mean of precision and recall. It is especially useful when the class distribution is imbalanced because it considers both false positives and false negatives.
\begin{equation*}
\text{F1 Score} = 2 \times \frac{\text{Precision} \times \text{Recall}}{\text{Precision} + \text{Recall}}
\end{equation*}
where Precision is $\frac{TP}{TP + FP}$ and Recall is $\frac{TP}{TP + FN}$.

For OOD dataset Twitch, we use the ROC-AUC as the evaluation metric. The ROC-AUC (Receiver Operating Characteristic - Area Under the Curve) is a performance measurement for classification problems. It evaluates the trade-off between the True Positive Rate (TPR) and False Positive Rate (FPR) at various threshold settings.

The AUC is calculated as:

\begin{equation*}
\text{AUC} = \int_{0}^{1} \text{TPR}(t) \, d\text{FPR}(t)
\end{equation*} 
Where:
\begin{itemize}
    \item \(\text{TPR}(t)\) is the True Positive Rate at threshold \(t\),
    \item \(\text{FPR}(t)\) is the False Positive Rate at threshold \(t\).
\end{itemize} 

\section{Comparative Analysis of Original and Reconnected Graph}\label{sec:reconnect}
During the training of node classification tasks, the labeled nodes are randomly sampled, which often results in these labeled nodes not being directly connected in the space. Consequently, the gradient defined in Equation~\ref{equ:gradient} cannot be computed. To address this, we propose two solutions, TAR (the method we use in the main text) and TAR-N, as detailed below:
\begin{itemize}
    \item \textbf{TAR.} This method sets the loss of the unlabeled nodes to the mean loss of the labeled nodes. This approach preserves the original graph structure, allowing the gradient flow between training nodes to occur indirectly through the unlabeled nodes.
    \item \textbf{TAR-N.} This method performs a breadth-first search for each labeled node during training to find $K$ reachable labeled nodes. Then, it adds edges between the current node and these $K$ nodes, creating a new graph, and performs the gradient flow on this new graph. This approach enables the gradient to propagate directly among the training nodes but disrupts the original graph structure.
\end{itemize}

We conduct experiments both on graph OOD datasets and class-imbalanced datasets to compare their performance. The results are summarized in Table~\ref{tab:ood_reconnect} and~\ref{tab:ci_reconnect}. TAR-N only achieved performance improvement over TAR on the Citerseer dataset and Twitch datasets under covariate shift, but this improvement was not significant. In other cases, TAR-N performed worse than TAR, which uses the original graph structure. This suggests that preserving the original graph structure might be more reliable. Additionally, TAR-N requires extra computational overhead for the breadth-first search compared to TAR.

\input{./data/table/ood_reconnect}
\input{./data/table/ci_reconnect}

\newpage
\section{Proof}\label{sec:proof}

\subsection{Proof of Theorem 1}
\begin{proof}
Denote $q^\star=\arg\sup\limits_{q\in\mathscr{P}_o(G_0)}\mathcal{L}(\theta,q)-\frac{1}{2\tau}\mathcal{GW}^2_{G_0}(p,q)$, and we have $\epsilon =\mathcal{GW}^2_{G_0}(p,q^\star)$.

Then we prove by contradiction:
we first assume $q^{'}=\arg\sup\limits_{q:\mathcal{GW}^2_{G_0}(p,q)\leq \epsilon}\mathcal{L}(\theta,q)$, which indicates that $\mathcal{L}(\theta,q^{'})\geq \mathcal{L}(\theta,q^\star)$ and $\mathcal{GW}^2_{G_0}(p, q^{'})\leq \epsilon$. 
Therefore, we have $\mathcal{GW}^2_{G_0}(p, q^{'})\leq \mathcal{GW}^2_{G_0}(p, q^{\star})$.
Denote $\mathcal{R}(\theta,q)=\mathcal{L}(\theta,q)-\frac{1}{2\tau}\mathcal{GW}_{G_0}^2(p,q)$, then we have $\mathcal{R}(\theta,q^\star)\leq \mathcal{R}(\theta,q^{'})$.
This leads to contradiction since $q^\star$ is the supremum point of $\mathcal{R}(\theta,\cdot)$.
\end{proof}

\subsection{Proof of Theorem 2}
\begin{proof}
	Based on the~\citep[Theorem 5]{chow2017entropy}, we have
\begin{equation*}
	\mathcal{L}(q(\infty)) - \mathcal{L}(q(t))\leq e^{-Ct}(\mathcal{L}(q(\infty))-\mathcal{L}(q(0))),
\end{equation*}
where $C>0$ is a constant depending on the loss function $\ell$, hyper-parameter $\beta$, and the sample size $N$.

Then we denote the real worst-case distribution within the $\epsilon$-radius discrete Geometric Wasserstein-ball as $q^*$, and we have
\begin{equation*}
	\mathcal{L}(q(\infty)) -\mathcal{L}(q^\star)+\mathcal{L}(p^\star) - \mathcal{L}(q(t))\leq e^{-Ct}(\mathcal{L}(q(\infty))-\mathcal{L}(q^\star)+\mathcal{L}(q^\star)-\mathcal{L}(q(0))).
\end{equation*}
Therefore, we have
\begin{equation*}
	\mathcal{L}(q^\star)-\mathcal{L}(q(t)) \leq e^{-Ct}(\mathcal{L}(q^\star)-\mathcal{L}(q(0)))-(1-e^{-Ct})(\mathcal{L}(q(\infty))-\mathcal{L}(q^\star)),
\end{equation*}
and
\begin{equation*}
	\frac{\mathcal{L}(q^\star)-\mathcal{L}(q(t))}{\mathcal{L}(q^\star)-\mathcal{L}(q(0))}\leq e^{-Ct} - (1-e^{-Ct})\frac{\mathcal{L}(q(\infty))-\mathcal{L}(q^\star)}{\mathcal{L}(q^\star)-\mathcal{L}(q(0))}<e^{-Ct}.
\end{equation*}
\end{proof}

\section{Limitations and Future Work}\label{sec:limitations}
In this paper, we propose a model-agnostic node classification OOD generalization algorithm TAR that leverages the existing topological information in node classification tasks to achieve better local robustness. However, as discussed in Appendix~\ref{sec:reconnect}, due to the random sampling of labeled nodes in node classification tasks, the nodes with computed losses are not directly connected, with other nodes without computed losses in between. Although we explored two potential solutions in Appendix~\ref{sec:reconnect}, we consider these solutions suboptimal. In the future, we aim to design a more effective method to address this issue.

%% file: data/table/ood_datasets.tex
\begin{table}[h]
    \centering
    \caption{Numbers of nodes in training, ID validation, ID test, OOD validation, and OOD test sets for the four datasets.}
    \label{tab:ood_datasets}
    \resizebox{0.8\textwidth}{!}{
    \begin{tabular}{c|cccccc}
    \toprule
        Dataset & Shift & Train & ID Validation &  ID test & OOD Validation & OOD Test\\ \midrule
        \multirow{2}{*}{WebKB} & concept & 282 & 60 & 60 & 106 & 109 \\
        ~ & covariate & 244 & 61 & 61 & 125 & 126 \\ \midrule
        \multirow{2}{*}{CBAS} & concept &  140 & 140 & 140 & 140 & 140 \\
        ~ & covariate & 420 & 70 & 70 & 70 & 70\\ \midrule
        \multirow{2}{*}{Twitch} & concept & 13605 & 2914 & 2914 & 6762 & 7925 \\
        ~ & covariate &  14448 & 3412 & 3412 & 6551 & 6297 \\ \midrule
        \multirow{2}{*}{Cora-Word} & concept & 7273 & 1558 & 1558 & 3807 & 5597 \\
        ~ & covariate & 9378 & 1979 & 1979 & 3003 & 3454 \\ 
        \bottomrule
    \end{tabular}}
\end{table}

%% file: data/table/ood_setting.tex
\begin{table}[h]
    \centering
    \caption{Architecture and hyperparameters on graph ood datasets}
    \label{tab:ood_setting}
    \resizebox{\textwidth}{!}{
    \begin{tabular}{c|cccccccc}
    \toprule
        Dataset & \multicolumn{2}{c}{CBAS} & \multicolumn{2}{c}{WebKB} & \multicolumn{2}{c}{Twitch} & \multicolumn{2}{c}{Cora} \\ \midrule
        Shift & concept & covariate & concept & covariate & concept & covariate & concept & covariate \\ \midrule
        Backbone & \multicolumn{8}{c}{Graph Convolutional Networks~\cite{gcn}} \\ \midrule
        Hidden dimension & \multicolumn{8}{c}{300} \\ 
        Number of layers & \multicolumn{8}{c}{3} \\ 
        Dropout & \multicolumn{8}{c}{0.5} \\ 
        Activation & \multicolumn{8}{c}{ReLU} \\ \midrule
        Epoch & \multicolumn{2}{c}{200} & \multicolumn{2}{c}{100} & \multicolumn{2}{c}{100} & \multicolumn{2}{c}{100} \\ 
        Batch size & \multicolumn{2}{c}{1000} & \multicolumn{2}{c}{4096} & \multicolumn{2}{c}{4096} & \multicolumn{2}{c}{4096} \\ 
        Backbone learning rate & \multicolumn{8}{c}{0.001} \\ \midrule
        TAR learning rate & 0.01 & 0.001 & 0.001 & 0.001 & 0.001 & 0.001 & 0.001 & 0.001 \\ 
        $T_{\text{in}}$ & 10 & 30 & 10 & 10 & 30 & 10 & 10 & 10 \\ 
        $\beta$ & 0.1 & 0.1 & 0.01 & 1 & 0.1 & 0.1 & 0.1 & 1 \\ \bottomrule
    \end{tabular}}
\end{table}

%% file: data/table/ci_datasets.tex
\begin{table}[h]
    \centering
    \caption{Label distribution of train class-imbalanced datasets [\%].}
    \label{tab:ci_datasets}
    \resizebox{0.7\textwidth}{!}{
    \begin{tabular}{c|ccccccc}
    \toprule
        Dataset & $\mathbf{L_0}$ & $\mathbf{L_1}$ & $\mathbf{L_2}$ & $\mathbf{L_3}$ & $\mathbf{L_4}$ & $\mathbf{L_5}$ & $\mathbf{L_6}$ \\ \midrule
        Cora-LT & 54.04 & 25.04 & 11.57 & 5.39 & 2.38 & 1.11 & 0.48 \\
        CiteSeer-LT & 60.72 & 24.06 & 9.49 & 3.76 & 1.47 & 0.49 & - \\
        PubMed-LT & 90.10 & 9.01 & 0.89 & - & - & - & - \\
    \bottomrule
    \end{tabular}}
\end{table}

%% file: data/table/ci_setting.tex
\begin{table}[t]
    \centering
    \caption{Architecture and hyperparameters on graph class-imbalanced datasets}
    \label{tab:ci_setting}
    \begin{tabular}{c|cccccc}
    \toprule
        \multirow{2}{*}{Dataset} & \multicolumn{2}{c}{Cora-LT} & \multicolumn{2}{c}{CiteSeer-LT} & \multicolumn{2}{c}{PubMed-LT} \\ 
        ~ & ERM & GraphENS & ERM & GraphENS & ERM & GraphENS \\ \midrule
        Backbone & \multicolumn{6}{c}{Graph Attention Networks\cite{gat}} \\ \midrule
        Hidden dimension & \multicolumn{6}{c}{256} \\
        Heads & \multicolumn{6}{c}{8} \\
        Number of layers & \multicolumn{6}{c}{2} \\
        Dropout & \multicolumn{6}{c}{0.5} \\
        Activation & \multicolumn{6}{c}{ELU} \\ \midrule
        Epoch & \multicolumn{6}{c}{500} \\
        Learning rate & \multicolumn{6}{c}{0.01} \\ \midrule
        TAR learning rate & 0.0001 & 0.0001 & 0.01 & 0.001 & 0.01 & 0.01 \\ 
        $T_{\text{in}}$ & 10 & 10 & 10 & 30 & 3 & 5 \\
        $\beta$ & 0.1 & 0.1 & 1 & 0.001 & 1 & 0.0001 \\ 
        \bottomrule
    \end{tabular}
\end{table}

%% file: data/table/ood_reconnect.tex
\begin{table}[h]
    \caption{The performance of TAR with orginal graph or TAR with reconnected graph on graph OOD datasets.}
    \vspace{0.1in}
    \label{tab:ood_reconnect}
    \centering
    \resizebox{0.9\textwidth}{!}{
    \begin{tabular}{c|cccccc}
    \toprule
        Dataset & \multicolumn{2}{c}{WebKB} & \multicolumn{2}{c}{CBAS} & \multicolumn{2}{c}{Twitch} \\
        Shift & concept & covariate & concept & covariate & concept & covariate \\ \midrule
        ERM & 26.97$\pm$1.49 & 14.13$\pm$2.92 & 82.86$\pm$1.28 & 78.43$\pm$1.00 & 47.87$\pm$0.65 & 48.55$\pm$0.91 \\ 
        TAR & \textbf{27.98$\pm$1.02} & \textbf{18.57$\pm$3.30} & \textbf{83.57$\pm$1.57} & \textbf{79.86$\pm$2.07} & \textbf{49.32$\pm$0.63} & 49.20$\pm$1.39 \\
        TAR-N & 27.89$\pm$0.94 & 16.27$\pm$3.73 & 83.36$\pm$0.72 & 78.71$\pm$1.62 & 49.22$\pm$0.90 & \textbf{49.22$\pm$1.48} \\
        \bottomrule
    \end{tabular}}
\end{table}

%% file: data/table/ci_reconnect.tex
\begin{table}[h]
    \centering
    \caption{The performance of TAR with orginal graph or TAR with reconnected graph on class-imbalanced datasets.}
    \vspace{0.1in}
    \label{tab:ci_reconnect}
    \resizebox{\textwidth}{!}{
    \begin{tabular}{l|ccc|ccc|ccc}
    \toprule
    \multicolumn{1}{c|}{\multirow{2}{*}{Method}} & \multicolumn{3}{c|}{Cora-LT}& \multicolumn{3}{c|}{CiteSeer-LT} & \multicolumn{3}{c}{PubMed-LT}\\  
                      & Acc & bAcc& F1 & Acc & bAcc& F1   & Acc  & bAcc & F1   \\ \midrule
    ERM & 73.04$\pm$0.15 & 63.83$\pm$0.32 & 63.67$\pm$0.60& 54.38$\pm$0.36 & 47.83$\pm$0.36   & 43.58$\pm$0.57 & 70.80$\pm$0.39 & 57.77$\pm$0.32 & 52.58$\pm$0.33 \\
    w/ TAR  & \textbf{74.17$\pm$0.38} & \textbf{66.10$\pm$0.76} & \textbf{66.29$\pm$0.58} & 57.08$\pm$0.53   & 50.41$\pm$0.56   & 47.01$\pm$0.77  & \textbf{75.40$\pm$0.43} & \textbf{67.33$\pm$0.59} & \textbf{68.23$\pm$0.75} \\
    w/ TAG-N & 73.64$\pm$0.22 & 65.20$\pm$0.41 & 66.12$\pm$0.42 & \textbf{57.50$\pm$0.44}   & \textbf{50.84$\pm$0.45}   & \textbf{47.73$\pm$0.61}  & 73.20$\pm$0.44 & 61.27$\pm$0.41 & 58.61$\pm$0.53\\
    \bottomrule
\end{tabular}}
\end{table}